\documentclass[runningheads]{llncs}
\usepackage[T1]{fontenc}
\usepackage{orcidlink}
\usepackage{bbding}
\usepackage{amsmath}
\usepackage{listings}
\usepackage[normalem]{ulem}
\usepackage{multirow}
\usepackage{array}
\usepackage{stfloats}
\usepackage{soul}
\usepackage{placeins}
\makeatletter
\let\@llncs@subparagraph\subparagraph
\renewcommand\subparagraph{\@startsection{subparagraph}{5}{\z@}%
  {3.25ex \@plus1ex \@minus .2ex}{-1em}%
  {\normalfont\normalsize\bfseries}}
\makeatother
\usepackage{titlesec}
\makeatletter
\let\subparagraph\@llncs@subparagraph  
\makeatother

\usepackage{tikz}
\usetikzlibrary{positioning,fit,calc,arrows.meta,decorations.pathreplacing}

\usepackage{graphicx}
\usepackage{color}

\titlespacing*{\subsubsection}{0pt}{0.6\baselineskip}{0.4\baselineskip}    

\usepackage{hyperref}
\usepackage{bookmark}

\bookmarksetup{
  depth=2,
  open=true,
}
\usepackage{fancyhdr}

\begin{document}
\title{Towards Lightweight Reliability: Using Soft Prompts for Hallucination Mitigation in Large Language Models}
\titlerunning{Lightweight Hallucination Mitigation Using Soft Prompts}

\author{S M Tahmid Siddiqui\inst{1}\textsuperscript{*(\Envelope)}\orcidlink{0009-0007-1931-0569}\and
Akib Jawad Ononto\inst{1}\thanks{These authors contributed equally to this work.} \orcidlink{0009-0007-2458-4695}\ \and
Anoop Singhal\inst{2}\orcidlink{0000-0002-2602-3927} \and
Latifur Khan\inst{1}\orcidlink{0000-0002-9300-1576}
}

\authorrunning{S. Siddiqui et al.}
%
\institute{The University of Texas at Dallas, Richardson, TX 75080, USA
\email{\{tahmid.siddiqui,akib,lkhan\}@utdallas.edu} 
\and
National Institute of Standards and Technology, Gaithersburg, USA
\email{anoop.singhal@nist.gov}
}
\maketitle              
\makeatletter\renewcommand{\@mkboth}[2]{}\makeatother
\renewcommand{\markright}[1]{}
\renewcommand{\markboth}[2]{}

\begin{tikzpicture}[remember picture,overlay]
\node[anchor=north west,align=center,text width=\textwidth,font=\footnotesize]
  at ([xshift=\dimexpr\oddsidemargin+1in\relax,yshift=-0.5cm]current page.north west)
  {Accepted for publication in DBSec 2026. The final publication is 
  
  available at Springer via \url{https://doi.org/10.1007/978-3-032-33260-8_14}.};
\end{tikzpicture}

\begin{abstract}
Large language models (LLMs) have seen widespread adoption across various domains, yet their reliability is frequently undermined by hallucinations - responses that are plausible-sounding but factually incorrect. In high-stakes domains, these errors can reduce trust and introduce real-world risk. To address this challenge, we present a parameter-efficient approach that uses soft prompts to mitigate hallucinated content and promote responsible abstention in generative question-answering (QA) tasks. 

Our method, called Responsible Contrastive Soft Prompting (RCSP), uses a composite loss to train soft prompts that balance three goals: suppressing hallucinatory content, encouraging abstention under uncertainty, and preserving or improving factual recall. To achieve these goals, we incorporate contrastive loss, curriculum learning, and KL regularization into our training mechanism. We evaluate our approach on five diverse generative QA datasets using an LLM-as-a-Judge framework. Experimental results on the Gemma 3 (12B) and Llama 3.1 (8B) backbones demonstrate that RCSP effectively balances factual recall with hallucination suppression and abstention, yielding a generally superior F-score over standard reasoning and instruction-based prompting baselines. Notably, these improvements are achieved by training only a fraction of the parameters required by other tuning techniques. Our results demonstrate that soft prompts provide a modular and computationally efficient path toward improving LLM reliability.


\end{abstract}

\section{Introduction}
Large language models (LLMs) have achieved widespread adoption for various applications in diverse domains, including agent-based workflows, chatbots and knowledge retrieval \cite{brown2020language,long2024generative,wu2025answer}. However, one primary challenge to the reliability of LLMs stems from hallucinations, i.e., responses that appear plausible but are factually incorrect or fabricated \cite{huang2025survey,kalai2025language}. Despite advancements in this domain, recent benchmarks such as HaluEval \cite{li2023halueval}, BLanCK \cite{chowdhury2025facts} and SimpleQA \cite{wei2024measuring} report that hallucinated content persists in a significant fraction of generated responses, which poses risks in high-stakes domains and undermines trust in deployed systems. 

Tackling the LLM hallucination challenge has remained a major point of focus in both academia and industry ever since the mass adoption of LLMs \cite{huang2025survey}, but responsible abstention is a relatively new objective for frontier LLMs \cite{kalai2025language,wei2024measuring,wu2025answer}. It entails teaching the model to acknowledge uncertainty to the best of its ability rather than guessing when uncertain in order to maximize accuracy. Instead of solely prioritizing higher benchmark accuracy, recent works advocate for a shift toward reducing hallucination rates. 

The rapid advancement of generative AI has also necessitated a shift in how systems are fine-tuned, evaluated and secured. As LLMs scale toward trillions of parameters, parameter-efficient techniques have attracted increasing interest for the purposes of improving model behavior \cite{houlsby2019parameter,prottasha2024parameter}. One such parameter-efficient technique is prompt tuning, or more informally, soft prompts \cite{lester2021power}. A soft prompt consists of a small number of learnable continuous prompt embeddings that are prepended to the input sequence while the backbone model remains frozen. Unlike traditional fine-tuning, which modifies the model's weights, soft prompting is a lightweight plug-and-play approach that allows for targeted behavioral shaping without requiring global distribution shifts.

In this paper, we introduce Responsible Contrastive Soft Prompting (RCSP), a training mechanism designed to promote responsible abstention and mitigate factual hallucinations in generative question-answering (QA). Our approach employs a composite loss function to train soft prompts that, when attached to a base LLM, nudge the LLM toward generating less hallucinatory content, abstaining when beneficial, and retaining or improving factual recall. Rather than modifying the full model parameters, which can be computationally expensive, we attach the lightweight soft prompts to frozen backbone LLMs. In addition, we incorporate a curriculum learning schedule and KL regularization within our contrastive objective to prevent the soft prompt from causing excessive abstention. We evaluate our approach using an abstention-aware LLM-as-a-Judge protocol that prioritizes reliability over simple accuracy metrics. By framing hallucination mitigation as an abstention-aware objective with explicit behavioral trade-offs, our method offers a lightweight and controllable pathway for improving LLM reliability in security-sensitive contexts.

\section{Motivation and Problem Statement}
LLMs are increasingly being integrated into security-sensitive applications, including knowledge retrieval, cyber forensics and malware detection \cite{ferrag2025generative}. In such settings, reliability is directly tied to system integrity and downstream risk. Hallucinated outputs can propagate misinformation, trigger incorrect automated actions or mislead human operators in fields like cybersecurity, potentially causing systems to ignore or even introduce vulnerabilities \cite{ferrag2025generative,huang2025survey}. In high-stakes environments, even a small fraction of unsupported claims can have disproportionate financial, operational, and legal consequences, including GDPR and HIPAA compliance breaches \cite{das2025security,kang2023deficiency}.

Importantly, recent findings highlight that hallucination is also a behavioral issue brought about by how models are trained and evaluated, rather than just being caused by a knowledge deficit \cite{kalai2025language}. Standard QA benchmarks for LLMs further reinforce this behavior since they grade abstentions the same as incorrect answers. Consequently, these benchmarks incentivize guessing since it is mathematically more reasonable to attempt an answer instead of abstaining, even in situations where the model lacks the required knowledge. On the other hand, methods that aggressively suppress hallucinations often degrade factual recall, while strategies that encourage abstention can make models overly cautious and reduce utility \cite{qin2026large}.

An ideally reliable system should therefore distinguish between answerable and unanswerable instances, providing factual responses when confident and abstaining when uncertainty is high. As Kalai et al. propose, the first step in changing the current landscape is to make evaluation mechanisms abstention-aware \cite{kalai2025language}. In addition, the mitigation approach should ideally be computationally efficient and modular to increase accessibility and adoption in real-world settings. 


To address this problem, we propose a Responsible Contrastive Soft Prompt (RCSP) framework that attaches learnable soft prompts to frozen backbone models. The `plug-and-play' nature of the implementation makes it easy to swap different soft prompts in and out if needed without modifying the underlying model, thus enabling greater customization and modularity. Our design also makes RCSP especially suitable for security-sensitive settings where responsible abstention and flexible implementations are vital, paving the way for safer and more trustworthy deployment of LLMs across domains. To systematically evaluate the effectiveness of RCSP, we investigate the following research questions:
\begin{itemize}
\item \textbf{RQ1} Can RCSP promote responsible abstention and reduce hallucination rates in generative QA without modifying backbone model parameters?
\item \textbf{RQ2} Can we utilize RCSP without causing degradation of factual recall and LLM utility?
\item \textbf{RQ3} Can soft prompts provide a computationally efficient and modular alternative to existing approaches for hallucination mitigation?
\end{itemize}

\section{Related Work}
Combating LLM hallucinations has become a key focus in both academia and industry, with research comprising two complementary pathways: detection and mitigation. In this paper, we work on hallucination mitigation, which is aimed at improving the factual reliability of generated responses. Hallucination mitigation in LLMs has been widely studied through a variety of approaches that work at different stages of the generation process. Existing methods differ in whether they steer model behavior through prompting, intervene during inference, rely on external knowledge sources or modify model parameters during training. In this section, we review prior work across prompting methods, inference-time interventions, retrieval-based grounding, fine-tuning approaches and preference-based alignment methods.

Prompt engineering has emerged as a lightweight alternative for influencing model behavior without retraining the underlying model. Several studies demonstrate that adding cautionary or deliberative instructions, such as encouraging careful reasoning or introducing emotional cues, can improve reliability \cite{li2023large,wang2023plan,wei2022chain}. Despite their simplicity, manual prompting strategies are highly sensitive to wording and lack tunable hyperparameters that allow for more granular tuning of the model's behavior. Their effectiveness varies across tasks, and they cannot adapt automatically to dataset-specific hallucination patterns.


Another line of work focuses on inference-time intervention techniques that aim to reduce hallucinations by steering model behavior during decoding without retraining or updating model parameters \cite{chuang2024dola,li2023inference}. Although these approaches avoid expensive training costs, they operate solely at inference time. As a result, they do not reshape the model’s learned objective or provide a mechanism for abstention-aware learning and they may potentially introduce additional latency.

One of the most widely adopted strategies for reducing hallucinations is retrieval-augmented generation (RAG) \cite{lewis2020retrieval}, which grounds model outputs in externally retrieved documents. By incorporating verified knowledge into consideration during response generation, retrieval-augmented systems can improve factual accuracy and reduce unsupported claims. However, these approaches require an external verified knowledge base and introduce latency during inference. Moreover, retrieval does not fundamentally alter the model’s internal generative behavior, so when relevant information is unavailable or retrieval fails, models may still hallucinate \cite{shuster2021retrieval}.

\begin{table}[ht]
\caption{Comparison of trainable parameter counts across selected fine-tuning approaches for the Gemma-3-12B-pt bf16 model.}
\label{tab:finetune-comparison}
\centering
\begin{tabular}{|l|l|r|r|}
\hline
\textbf{Method} & \textbf{Configuration} & \textbf{Trainable Parameters} & \textbf{\% of Parameters} \\
\hline
R-Tuning \cite{zhang2024rtuning} & Full fine-tuning & 12{,}335{,}656{,}560 & 100.00\% \\
QLoRA \cite{dettmers2023qlora} & Rank 32 (all linear layers) & 148{,}331{,}520 & $\sim$1.202\% \\
RCSP & 40 tokens (input layer) & 1{,}157{,}120 & \textbf{$\sim$0.009\%} \\
\hline
\end{tabular}
\end{table}

A large body of work attempts to mitigate hallucinations by modifying model parameters through supervised fine-tuning \cite{devlin2019bert,ouyang2022training}. Traditional full fine-tuning updates all parameters of a pretrained model using supervised task-specific data, enabling strong adaptation to downstream tasks. Recent work such as Refusal-Aware Instruction Tuning (R-Tuning) \cite{zhang2024rtuning} trains models to recognize knowledge gaps by encouraging abstention responses such as ``I don't know'' when questions are unanswerable. However, for modern LLMs with billions of parameters, fine-tuning approaches require substantial computational resources and may introduce risks such as catastrophic forgetting, where previously learned general capabilities degrade during specialization \cite{kirkpatrick2017overcoming,luo2025forgetting}.

To address these limitations, parameter-efficient fine-tuning (PEFT) methods have been proposed to adapt LLMs while updating a small subset of parameters. Adapter-based approaches introduce lightweight task-specific modules within transformer layers \cite{houlsby2019parameter}, while Low-Rank Adaptation (LoRA) \cite{hu2022lora} learns low-rank update matrices that approximate weight updates without modifying the full parameter space. QLoRA \cite{dettmers2023qlora} further improves efficiency by combining low-rank adaptation with quantized model weights, enabling large models to be fine-tuned with lower memory requirements. While these methods significantly reduce training cost, they still introduce trainable parameters within the model's internal layers. As shown in Table \ref{tab:finetune-comparison}, RCSP requires orders of magnitude fewer trainable parameters than variants of fine-tuning approaches, which typically rely on standard supervised or preference-based objectives rather than explicitly targeting hallucination suppression. As a result, such approaches may reduce modularity and complicate controlled behavioral interventions in safety-critical settings.

Preference-based alignment methods also modify model parameters through supervised fine-tuning or preference optimization. Instruction tuning and reinforcement learning from human feedback (RLHF) \cite{ouyang2022training} align models with human preferences, including encouraging truthful responses. More recently, methods such as Direct Preference Optimization (DPO) \cite{rafailov2023direct} train models to align with human preference pairs through a simplified optimization objective. While these approaches improve factual reliability, they require updating model parameters and often depend on large-scale preference datasets that are frequently infeasible to obtain.

Soft prompt methods provide a parameter-efficient alternative by steering model behavior through learnable prompts while keeping the backbone model frozen \cite{lester2021power}. For example, Soft Prompting for Unlearning (SPUL) \cite{bhaila2025soft} learns task-specific soft prompts specifically designed for the unlearning domain to remove unwanted knowledge from pretrained models. Such approaches have demonstrated that behavioral steering can be achieved through lightweight prompt representations; however, they are not specifically designed to address hallucination mitigation and abstention-aware behavior.

\section{Methodology}
\begin{figure}[t]
\centering
\resizebox{\columnwidth}{!}{%
\begin{tikzpicture}[
    font=\normalsize,
    >=Stealth,
    line/.style={->, line width=0.85pt},
    box/.style={
        draw,
        rounded corners=6pt,
        minimum height=1.2cm,
        align=center,
        line width=0.75pt,
        inner sep=1.2pt
    }
]

\fill[blue!4, rounded corners=4pt] (-1.0,-0.95) rectangle (12.95,2.45);
\fill[green!6, rounded corners=4pt] (-1.0,-3.25) rectangle (12.95,-1.35);
\node[font=\small\bfseries, anchor=west] at (-0.8,2.2) {Phase 1: Offline Soft Prompt Training \& Selection (once)};
\node[font=\small\bfseries, anchor=west] at (-0.8,-1.60) {Phase 2: Deployment-time Inference};

\node[box, fill=blue!12, text width=2.55cm] (contrast) at (0.7,1.35)
{Contrastive dataset\\$(q,h,f)$};
\node[box, fill=blue!12, text width=2.55cm] (utility) at (0.7,-0.15)
{Utility dataset\\$(q,f)$};
\node[box, sharp corners, fill=orange!15, text width=3.1cm] (loss) at (4.35,0.6)
{Composite Loss Optimization\\
$\mathcal{L}_{contrast}+\mathcal{L}_{utility}$\\};
\node[box, fill=green!20, text width=2.95cm] (adapter) at (4.3,-2.45)
{Selected Soft Prompt};

\draw[line] (contrast.east) -- (loss.west);
\draw[line] (utility.east) -- (loss.west);

\node[box, fill=green!14, text width=2.95cm] (saved) at (8.15,0.6)
{Learned Soft Prompts};
\node[box, sharp corners, fill=cyan!12, text width=2.6cm] (eval) at (11.55,0.6)
{LLM-as-a-Judge Evaluation};

\draw[line] (loss.east) -- (saved.west);
\draw[line] (saved.east) -- (eval.west);
\draw[->, densely dotted, line width=0.85pt] (eval.south west) to[out=-130, in=20] (adapter.north east);

\node[box, fill=gray!10, text width=2.55cm] (question) at (0.7,-2.45)
{User question};
\node[font=\normalsize] (merge) at ($(question.east)!0.5!(adapter.west)$) {$+$};
\node[box, sharp corners, fill=yellow!20, text width=2.95cm] (llm) at (8.45,-2.45)
{Frozen backbone LLM (e.g. Gemma-3-12B)};
\node[box, fill=purple!12, text width=2.5cm] (answer) at (11.85,-2.45)
{Generated answer};

\draw[line] (llm.east) -- (answer.west);
\draw[line] ([xshift=0.35cm]adapter.east) -- (llm.west);
\draw[decorate, decoration={brace, mirror, amplitude=4.5pt}, line width=0.7pt]
  ([yshift=-0.33cm]question.south west) -- ([yshift=-0.33cm]adapter.south east)
  node[midway, yshift=-0.50cm] {};
\end{tikzpicture}%
}
\caption{\textbf{Two-phase RCSP workflow}. Phase 1 (offline) trains soft prompts using a composite loss function, then evaluates the learned soft prompts using a judge LLM to identify the best-performing one. Phase 2 (deployment) prepends the soft prompt selected in phase 1 with the user question and feeds it to the base LLM to generate the final answer. Phase 1 is performed once prior to deployment while Phase 2 executes for each user query.}
\label{fig:pipeline}
\end{figure}
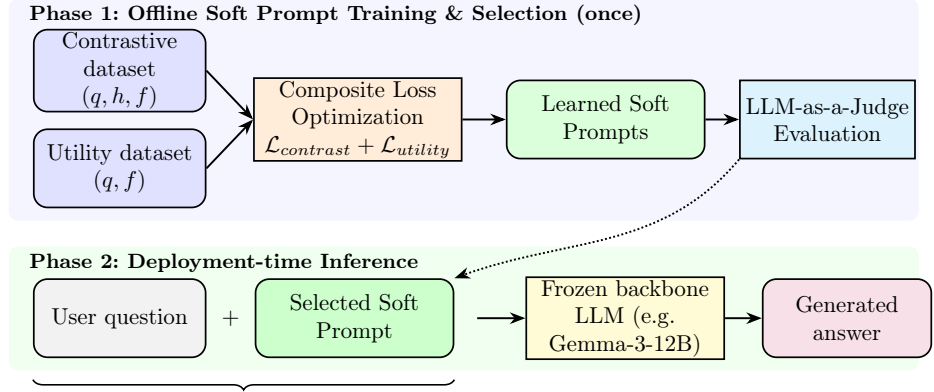

\subsection{Hallucination Mitigation Pipeline Overview}
Before detailing the specific training objectives, we present the high-level pipeline of our proposed framework, illustrated in Figure~\ref{fig:pipeline}. Our framework is strictly divided into two distinct operational phases.

\textbf{Phase 1: Offline Soft Prompt Training and Selection} The upper region of Figure~\ref{fig:pipeline} depicts the one-time training phase. The soft prompts are optimized using a composite loss function over both the contrastive dataset and the utility dataset. Since soft prompt training requires direct access to the model's embedding layer, we use open-weight models for our approach. The backbone LLM remains frozen throughout this phase. Once the training is complete, the learned soft prompts are saved and evaluated on validation subsets using a judge LLM to select the best-performing soft prompt configuration for downstream evaluation.

\textbf{Phase 2: Deployment-time Inference} The lower region of Figure~\ref{fig:pipeline} illustrates the per-query deployment phase. For every new user question, the soft prompt identified during Phase 1 is prepended to the input question. No gradient updates occur during inference. The frozen LLM then generates a response conditioned on this combined prompt.


\subsection{Soft Prompt Training Mechanism}
At the core of our approach is our implementation of the composite loss function, defined in Equation~\eqref{composite}. The objective of the composite loss function is to train soft prompts that simultaneously reduce hallucinated responses while preserving factual knowledge and promoting responsible abstention. 
\begin{equation}
\label{composite}
\mathcal{L} = \mathcal{L}_{\text{contrast}} + \mathcal{L}_{\text{utility}}
\end{equation}
This composite objective explicitly separates hallucination avoidance and knowledge preservation. Unlike regular fine-tuning objectives that treat all supervision uniformly, our setup grants greater control over targeted suppression and the retention of general model utility.
\subsubsection{The Contrastive Loss} 
One half of the composite loss objective is the contrastive loss, which is denoted as $L_{\text{contrast}}$ in Equation~\eqref{composite}. The contrastive loss is applied to samples drawn from the hallucination suppression dataset $D_{\text{contrast}}$, which consists of $(q,h,f)$ triplets where $q$ is the question, $h$ is a hallucinated response, and $f$ is the verified factual response. 

The contrastive loss penalizes the soft prompt for causing the model to align with hallucinated targets from $D_{\text{contrast}}$ and incorporates a curriculum learning schedule that gradually shifts the soft prompt's focus from abstention to factual answering during training. The curriculum learning schedule helps prevent the soft prompt from making the model overly prone to abstaining, which would degrade its factual recall performance. It also incorporates Kullback-Leibler (KL) regularization over the answer token span to prevent large deviations from the base model's output due to the addition of the contrastive loss. KL regularization encourages the soft prompt to behave as a perturbation of the base model rather than a full behavioral rewrite.  We define our contrastive loss in Equation~\eqref{contrast}.
\begin{equation}
\label{contrast}
\mathcal{L}_{\text{contrast}} = (1 - \mu)\, \mathrm{CE}(a) + \mu\, \mathrm{CE}(f) - \lambda\, \mathrm{CE}(h) + \beta_{\text{contrast}}\, \mathrm{KL}(p_{\theta} \parallel p_{\text{ref}})
\end{equation}
Here, $\mu$ controls the curriculum interpolation between abstention-focused and factual-answer supervision by adjusting the relative weighting of the corresponding cross-entropy terms. From Equation~\eqref{contrast}, we see that a smaller $\mu$ emphasizes abstention targets $a$ (e.g., “I don’t know”), while a larger $\mu$ increases supervision toward factual targets $f$. Initially, $\mu$ is set to a $\mu_{start}$ value. Over the training interval $T_{start}$ to $T_{end}$, the variable $\mu$ linearly increases toward $\mu_{end}$, where $\mu_{end} \geq \mu_{start}$. This transition shifts supervision toward factual-answering reinforcement as training progresses. This curriculum encourages the model to first learn to abstain from hallucinatory answers before refining its ability to provide factual answers when possible. The ideal outcome with this approach is that the model will learn calibrated abstention behavior without sacrificing factual recall. 

$\lambda$ controls the penalty applied to the hallucinated response, and $\beta_{\text{contrast}}$ scales the KL-regularization. $\mathrm{KL}(p_{\theta} \parallel p_{\text{ref}})$ denotes the Kullback--Leibler divergence between the current model distribution $p_{\theta}$ (conditioned on the learned soft prompt) and the base model distribution $p_{\text{ref}}$. 

We use $\mathrm{CE}(\cdot)$ to denote the cross-entropy loss for the corresponding target (abstention $a$, factual answer $f$, or hallucinated answer $h$). 
The negative cross entropy term in Equation~\eqref{contrast} discourages likelihood assignment to hallucinated targets, thus acting as a contrastive anti-supervision signal. We compute token-level cross entropy over answer spans only. For a target sequence $y$, we define the token-level cross entropy loss as:

\begin{equation}
\label{eq:cross-entropy}
\mathrm{CE}(y) = -\frac{1}{|M|} \sum_{t \in M} \log p_{\theta}(y_t \mid y_{<t}, q)
\end{equation}
where $M$ denotes the set of valid answer tokens (non-masked positions). Normalization by $|M|$ prevents undue penalization of longer answers which would otherwise accumulate larger absolute loss due to containing more tokens. This formulation ensures supervision is applied only to answer tokens and not to prompt tokens. 

\subsubsection{The Utility Loss}
The second component of the composite objective in Equation~\eqref{composite} is the utility loss $L_{\text{utility}}$, which aims to maintain factual question-answering ability. The utility loss is applied to $(q, f)$ pairs drawn from the utility dataset $D_{\text{utility}}$. We define the utility loss as follows:
\begin{equation}
\label{eq:utility-loss}
\mathcal{L}_{\text{utility}} = \mathrm{CE}(f) + \beta_{\text{utility}}\, \mathrm{KL}(p_{\theta} \parallel p_{\text{ref}})
\end{equation}
This mirrors the contrastive loss structure by using similar notation but excludes anti-supervision and curriculum components. The cross-entropy term preserves factual generation ability while the KL regularization constrains deviation from the base model distribution, thus helping prevent catastrophic forgetting.

\section{Experiments}
\subsection{Datasets}
To facilitate the training of our soft prompts, we partition our training corpus into two distinct sets - the contrastive and utility sets. As the name suggests, the contrastive part of the composite loss function utilizes the contrastive set to explicitly encourage reliable responses while suppressing hallucinated ones. The set consists of the first 8,000 questions from HaluEval \cite{li2023halueval} and the first 600 questions from TruthfulQA \cite{lin2022truthfulqa}. Similarly, the utility part of the composite loss function makes use of the utility set to preserve the model’s general factual answering capability. It consists of a set of 27,000 factual question–answer pairs drawn from TriviaQA \cite{joshi2017triviaqa}. 

For evaluation, we use five QA datasets: HaluEval \cite{li2023halueval}, TruthfulQA \cite{lin2022truthfulqa}, SimpleQA \cite{wei2024measuring}, MedHallu \cite{pandit2025medhallu} and WebQuestions \cite{berant2013semantic}. We use held-out test subsets of TruthfulQA and HaluEval to ensure that evaluation questions are excluded from the training data to prevent leakage. These datasets cover a diverse range of domains and question styles, allowing us to assess model behavior across different settings. To ensure a consistent and computationally efficient evaluation, we evaluate on a fixed subset of 500 questions sampled from each dataset across all methods and hyperparameter runs, except for TruthfulQA, where the evaluation split contains 217 questions. 
\subsubsection{HaluEval}
The HaluEval benchmark is a collection of generated and human-annotated samples specifically designed to evaluate and analyze hallucinations in LLMs. For our approach, we focus specifically on the samples from the question-answering (QA) task within this benchmark. Its hallucinated samples are generated via a two-stage sampling-then-filtering framework. Its inclusion of plausible and difficult hallucinated responses in addition to factual answers makes it uniquely suited for contrastive objectives. In addition, HaluEval aims to cover four distinct hallucination patterns, namely comprehension, factualness, specificity, and inference \cite{zheng2023does}, which in turn increases the breadth of our experiments as well. 
\subsubsection{TruthfulQA}
TruthfulQA is a benchmark designed to measure whether language models generate truthful answers to questions that commonly trigger misconceptions. The dataset contains adversarially designed questions spanning 38 categories such as health, law, finance and politics. For our training, we extract a tuple consisting of the question, the best incorrect answer and the best correct answer to form our ($q,h,f)$ tuple. The adversarial misconception-driven design of TruthfulQA introduces additional challenging hallucination patterns.

\subsubsection{TriviaQA} 
TriviaQA is a large-scale trivia QA dataset authored by trivia enthusiasts. The dataset requires reasoning across multiple sentences and handling lexical and syntactic variation. For our approach, we incorporate factual QA pairs from TriviaQA into the utility training set, where each sample consists of a question and its verified answer. 

\subsubsection{SimpleQA}
SimpleQA is a factual QA benchmark spanning a diverse array of topics. It consists of questions paired with verified reference answers. It is a relatively new and challenging benchmark designed to stress-test factual accuracy from frontier models like GPT-4. 

\subsubsection{MedHallu}
MedHallu is a medical hallucination benchmark consisting of QA pairs derived from PubMedQA \cite{jin2019pubmedqa}, a biomedical question-answering dataset based on PubMed literature. As hallucinated medical information can pose direct safety risks, this benchmark provides a high-stakes evaluation setting for hallucination suppression.
\subsubsection{WebQuestions} 
WebQuestions is an open-domain QA benchmark consisting of natural language questions collected from real user queries using the Google Suggest API and paired with verified answers grounded in the Freebase knowledge base. The dataset contains factoid questions about real-world entities and relationships, such as people, locations, dates and organizations. In our evaluations, WebQuestions provides a realistic evaluation of hallucination suppression in natural information-seeking scenarios.

\subsection{Evaluation}
Evaluating open-ended generation poses unique challenges \cite{wang2024learning}. Exact string matching is often insufficient for short-answer question answering, as semantically correct responses may differ lexically from reference answers. To address this, recent work has introduced LLM-as-a-Judge protocols \cite{zheng2023judging}, where a separate language model evaluates generated responses according to a structured rubric.

We evaluate the proposed soft prompt training approach using a generation-based question-answering evaluation framework combined with an LLM-based semantic grading protocol. Evaluations are conducted on multiple QA datasets consisting of brief question–answer pairs and the best-performing soft prompt and its hyperparameter configuration are identified. For each evaluation instance, the model generates a short response to a given prompt. The generated response is then evaluated by the judge model (OpenAI GPT-4.1 \cite{openai2025gpt41}), which is provided with the correct answer and a semantic evaluation rubric. For rigor and reliability, we adopt the rubric provided in the companion paper for the SimpleQA benchmark \cite{wei2024measuring} and modify it slightly to enforce stricter classification. Specifically, our modified grader template treats degenerate answers such as off-topic or question-echoing responses as hallucinations rather than ‘not attempted’ (see Appendix \ref{app:hyperparameters}). We use deterministic decoding during the generation process with temperature, top-p and beam size values set to 0, 1 and 1, respectively. 

Generated outputs are post-processed to remove trailing artifacts, repeated prompts, and multi-paragraph continuations before being fed to the judge. All evaluations are performed in inference mode without gradient updates. When soft prompt adapters are attached to the model, it is done so using the PEFT library. The judge model assigns one of three labels to each generated response:
\begin{itemize}
\item FACT: The response contains all required factual information and does not contradict the reference answer.
\item HALL: The response contradicts the reference answer or contains unsupported factual claims.
\item ABS: The response is an abstention, i.e. explicitly expresses uncertainty (e.g., ``I don't know'') and does not provide contradictory information.
\end{itemize}

\subsection{Baselines}
Our primary objective is to investigate whether our soft prompts can improve the reliability of a strong base LLM, similar to the goal stated in the Self-Refine project \cite{madaan2023self}. We select prompt-based baselines because they are most comparable to RCSP in terms of parameter efficiency, memory footprint and stage of intervention in the generation pipeline. Unlike fine-tuning or inference-time modification approaches, these methods all operate by modifying the input prompt without altering model weights or decoding dynamics. This ensures a controlled comparison within the class of lightweight plug-and-play interventions.

\subsubsection{Base} We use \textit{base} to refer to the original pretrained LLM available for use through the HuggingFace library.
\subsubsection{Append-idk} Instead of trainable vectors, we investigate the effect of appending English instructions that describe the desired behavior. The approach denoted as \textit{append-idk} refers to appending this string to our questions: ``Reply `I don't know' if you don't know the answer''.
\subsubsection{EmoPrompt} Li et al. \cite{li2023large} explored the effect of appending emotionally framed phrases derived from psychological theories to prompts. The paper identified 11 emotional stimuli that were tested across multiple LLMs and demonstrated that emotional language could be leveraged to improve the performance of LLMs, including transferability to tasks outside of the testing domains. Following the average metric the authors report in their paper, we also average the performance of all 11 stimuli prompts on each of our test datasets for our results.

\subsubsection{Plan-and-Solve} Plan-and-Solve \cite{wang2023plan} is a reasoning-based prompting approach that extends the seminal zero-shot Chain-of-Thought paper \cite{kojima2022large} by instructing the LLM to devise a plan to tackle the task at hand and then solve it according to the plan step-by-step. The method encourages the LLM to identify subtasks and pay close attention to the required calculations and intermediate steps to induce greater accuracy. We include it as a baseline to assess whether reasoning-based prompting can improve reliability and reduce hallucinations.

\subsection{Metrics}
A key challenge in hallucination mitigation is avoiding over-refusal, where a model becomes so cautious that it fails to provide factual information. At first glance, comparing the threefold tradeoffs based on the factual, abstention and hallucination percentages can be unintuitive in many cases. To better illustrate this trade-off and provide a unified metric, we report an F-score similar to the one proposed by the SimpleQA benchmark \cite{wei2024measuring} in addition to the percentages of abstentions, hallucinations and factual responses. We record the overall percentage of questions answered correctly, which we denote as Overall Accuracy ($A$), and compute the percentage of questions answered correctly out of those that the model chose to attempt, which we denote as Conditional Accuracy ($C$). These are defined as:
\[
A = \frac{c}{c + i + a}, \quad C = \frac{c}{c + i}
\]
\noindent where c, i and a represent the number of correct, incorrect and abstained responses, respectively. Finally, we compute the F-score as the harmonic mean of $A$ and $C$ as demonstrated in the equation below.
\[
F\text{-score} = 2 \cdot \frac{A \cdot C}{A + C} = \frac{2c}{2c + 2i + a}
\]

\section{Results}
\subsection{Primary Results}

We evaluated the generative QA performance using the Gemma 3 (12B) \cite{gemma3technicalreport2025} and the Llama 3.1 (8B) \cite{grattafiori2024llama} backbones across five diverse datasets to compare our RCSP method against the selected baselines. We selected models of different parameter sizes from two distinct families to assess whether RCSP generalizes across different architectures. All evaluations were conducted under a unified decoding setup and were run on L4, A100 or H100 GPUs depending on availability and memory requirements. The hyperparameter configurations used to train the reported soft prompts are provided in Table \ref{tab:hyperparams} inside the Appendix. The results of our evaluation process using Gemma 3 (12B) and Llama 3.1 (8B) are summarized in Tables~\ref{tab:gemma-qa-evals} and \ref{tab:llama-qa-evals} respectively.


\begin{table*}[!ht]
\caption{Generative QA evaluations for Gemma-3-12b-pt. ``HALL'', ``ABS'', and ``FACT'' denote percentages (\%) of hallucinated responses, abstentions, and factual responses as determined by the judge. Best F-score (highest) within each dataset is \textbf{bolded}.}
\label{tab:gemma-qa-evals}
\centering
\setlength{\tabcolsep}{6pt}
\renewcommand{\arraystretch}{1.1}
\small
\begin{tabular}{>{\raggedright\arraybackslash}p{2.5cm} l c c c c}
\hline
\textbf{Dataset} & \textbf{Method} & \textbf{HALL} $\downarrow$ & \textbf{ABS} & \textbf{FACT} $\uparrow$ & \textbf{F-score} $\uparrow$ \\
\hline
\multirow{5}{=}{HaluEval (test)} & Base & 58.00 & 4.00 & 38.00 & 38.77 \\
 & Append-idk & 54.67 & 9.33 & 36.00 & 37.76 \\
 & EmoPrompt (avg) & 55.79 & 8.73 & 35.48 & 37.10 \\
 & Plan-and-Solve & 56.67 & 5.67 & 37.67 & 38.77 \\
 & \underline{RCSP} & 55.00 & 6.00 & 39.00 & \textbf{40.21} \\
\cline{1-6}
\multirow{5}{=}{TruthfulQA (test)} & Base & 67.74 & 1.38 & 30.88 & 31.09 \\
 & Append-idk & 69.59 & 7.37 & 23.04 & 23.92 \\
 & EmoPrompt (avg) & 67.28 & 5.61 & 27.11 & 27.89 \\
 & Plan-and-Solve & 72.67 & 12.67 & 14.67 & 15.66 \\
 & \underline{RCSP} & 65.90 & 2.77 & 31.33 & \textbf{31.77} \\
\cline{1-6}
\multirow{5}{=}{WebQuestions} & Base & 40.00 & 5.00 & 55.00 & 56.41 \\
 & Append-idk & 39.67 & 6.33 & 54.00 & 55.77 \\
 & EmoPrompt (avg) & 39.58 & 7.91 & 52.51 & 54.68 \\
 & Plan-and-Solve & 45.33 & 2.67 & 52.00 & 52.70 \\
 & \underline{RCSP} & 37.33 & 6.67 & 56.00 & \textbf{57.93} \\
\cline{1-6}
\multirow{5}{=}{MedHallu} & Base & 62.00 & 9.00 & 29.00 & 30.37 \\
 & Append-idk & 66.67 & 6.67 & 26.66 & 27.58 \\
 & EmoPrompt (avg) & 59.64 & 16.06 & 24.30 & 26.43 \\
 & Plan-and-Solve & 61.67 & 24.67 & 13.67 & 15.59 \\
 & \underline{RCSP} & 57.67 & 12.67 & 29.67 & \textbf{31.67} \\
\cline{1-6}
\multirow{5}{=}{SimpleQA} & Base & 87.20 & 3.80 & 9.00 & 9.17 \\
 & Append-idk & 85.00 & 6.67 & 8.33 & 8.62 \\
 & EmoPrompt (avg) & 82.52 & 9.64 & 7.84 & 8.24 \\
 & Plan-and-Solve & 80.67 & 10.33 & 9.00 & 9.49 \\
 & \underline{RCSP} & 86.33 & 3.33 & 10.33 & \textbf{10.51} \\
\hline
\end{tabular}
\end{table*}

\begin{table*}[!t]
\caption{Generative QA evaluations for Llama 3.1-8b. ``HALL'', ``ABS'', and ``FACT'' denote percentages (\%) of hallucinated responses, abstentions, and factual responses as determined by the judge. Best F-score (highest) within each dataset is \textbf{bolded}.}
\label{tab:llama-qa-evals}
\centering
\setlength{\tabcolsep}{6pt}
\renewcommand{\arraystretch}{1.1}
\small
\begin{tabular}{>{\raggedright\arraybackslash}p{2.5cm} l c c c c}
\hline
\textbf{Dataset} & \textbf{Method} & \textbf{HALL} $\downarrow$ & \textbf{ABS} & \textbf{FACT} $\uparrow$ & \textbf{F-score} $\uparrow$ \\
\hline
\multirow{5}{=}{HaluEval (test)} & Base & 59.67 & 6.33 & 34.00 & 35.11 \\
 & Append-idk & 60.00 & 4.33 & 35.67 & 36.46 \\
 & EmoPrompt (avg) & 63.15 & 6.67 & 30.18 & 31.25 \\
 & Plan-and-Solve & 50.33 & 17.67 & 32.00 & 35.10 \\
 & \underline{RCSP} & 60.67 & 1.33 & 38.00 & \textbf{38.26} \\
\cline{1-6}
\multirow{5}{=}{TruthfulQA (test)} & Base & 76.50 & 1.38 & 22.12 & 22.27 \\
 & Append-idk & 73.27 & 5.99 & 20.74 & 21.38 \\
 & EmoPrompt (avg) & 75.66 & 3.23 & 21.11 & 21.46 \\
 & Plan-and-Solve & 82.33 & 5.00 & 12.67 & 12.99 \\
 & \underline{RCSP} & 71.43 & 3.69 & 24.88 & \textbf{25.35} \\
\cline{1-6}
\multirow{5}{=}{WebQuestions} & Base & 52.00 & 3.00 & 45.00 & 45.69 \\
 & Append-idk & 48.00 & 6.33 & 45.67 & 47.16 \\
 & EmoPrompt (avg) & 46.03 & 7.82 & 46.15 & 48.00 \\
 & Plan-and-Solve & 40.67 & 9.00 & 50.33 & 52.71 \\
 & \underline{RCSP} & 41.33 & 7.00 & 51.67 & \textbf{53.50} \\
\cline{1-6}
\multirow{5}{=}{MedHallu} & Base & 67.33 & 10.67 & 22.00 & 23.24 \\
 & Append-idk & 47.33 & 37.33 & 15.33 & 18.85 \\
 & EmoPrompt (avg) & 60.97 & 19.36 & 19.67 & 21.66 \\
 & Plan-and-Solve & 65.00 & 7.33 & 27.67 & \textbf{28.72} \\
 & \underline{RCSP} & 64.33 & 16.67 & 19.00 & 20.73 \\
\cline{1-6}
\multirow{5}{=}{SimpleQA} & Base & 90.67 & 2.67 & 6.67 & 6.76 \\
 & Append-idk & 91.00 & 3.33 & 5.67 & 5.76 \\
 & EmoPrompt (avg) & 91.00 & 3.73 & 5.27 & 5.38 \\
 & Plan-and-Solve & 86.33 & 9.00 & 4.67 & 4.89 \\
 & \underline{RCSP} & 92.00 & 0.67 & 7.33 & \textbf{7.35} \\
 \hline
\end{tabular}
\end{table*}

Overall, RCSP achieves the highest F-score on the majority of datasets across both Gemma-3 (12B) and Llama 3.1 (8B), indicating a consistently favorable balance between hallucination suppression, abstention behavior and factual accuracy. Across backbones and model sizes, RCSP maintains or improves factual response rates relative to the base model while reducing hallucination rates in most cases.

A key trend observed across both tables is that instruction-based prompting methods introduce noticeable trade-offs. While approaches such as Append-idk and EmoPrompt can reduce hallucinations in certain settings, they often do so at the cost of reduced factual recall or increased abstention. Similarly, reasoning-based prompting via Plan-and-Solve does not consistently improve reliability, and in several cases leads to substantial degradation in factual performance, particularly on more challenging datasets. The one dataset on which RCSP does not achieve the top F-score is MedHallu with Llama 3.1 (8B). By contrast, on the larger Gemma 3 (12B) backbone, RCSP performs well on MedHallu, achieving both the lowest hallucination rate and the highest F-score. This suggests that the soft prompt's contrastive training signal may interact differently with the Llama-3.1 architecture in the presence of domain-specialized medical content.

Another consistent finding across both models is that SimpleQA remains a particularly challenging benchmark, with hallucination rates exceeding 80\% for all methods regardless of backbone. This aligns with the benchmark's design intent as a frontier-model stress test. Despite the difficulty of the task, RCSP achieves the highest factual response rate and F-score on SimpleQA for both backbones, suggesting that even marginal gains in reliability on this benchmark are meaningful and non-trivial to obtain through lightweight interventions. Altogether, RCSP's demonstration of more stable and balanced behavior across datasets and model families relative to the other baselines points to the efficacy of RCSP's more nuanced mitigation strategy.


%

\subsection{Ablation Studies}
Ablation experiments with the Gemma-3 (12B) backbone were conducted on TruthfulQA to analyze the contribution of each component of our composite objective. Each ablated variant was trained from scratch with the corresponding loss component removed while all other training settings were kept identical. We compare RCSP against the baselines and targeted removals of individual loss components. The results are summarized in Table~\ref{tab:ablation-truthfulqa-full}.
\begin{table}[!htbp]
\caption{Ablation results for Gemma-3-12b-pt on TruthfulQA. ``HALL'', ``ABS'', and ``FACT'' denote percentages of judge-assigned hallucinated responses, abstentions, and factual responses (\%). Best results for HALL (lowest) and F-score (highest) are \textbf{bolded}.}
\label{tab:ablation-truthfulqa-full}
\centering
\setlength{\tabcolsep}{3pt}
\renewcommand{\arraystretch}{1.05}
\resizebox{\linewidth}{!}{%
\begin{tabular}{l|ccc|c}
\hline
\textbf{Method} & \textbf{HALL} $\downarrow$ & \textbf{ABS} & \textbf{FACT} $\uparrow$ & \textbf{F-score} $\uparrow$ \\
\hline
Base & 67.74 & 1.38 & 30.88 & 31.09 \\
Append-idk & 69.59 & 7.37 & 23.04 & 23.92 \\
EmoPrompt (avg) & 67.28 & 5.61 & 27.11 & 27.89 \\
Plan-and-Solve & 72.67 & 12.67 & 14.67 & 15.66 \\
\hline
\underline{RCSP} & \textbf{65.90} & 2.77 & 31.33 & \textbf{31.77} \\
\hline
w/o curriculum (fixed $\mu$) & 71.43 & 1.84 & 26.73 & 26.98 \\
w/o KL regularization ($\beta_{\text{contrast}}=\beta_{\text{utility}}=0$) & 69.59 & 2.76 & 27.65 & 28.04 \\
w/o $\mathcal{L}_{\text{utility}}$ & 67.28 & 1.84 & 30.88 & 31.17 \\
w/o $\mathcal{L}_{\text{contrast}}$ & 70.05 & 1.84 & 28.11 & 28.37 \\
\hline
\end{tabular}
}%
\end{table}

\begin{figure}[!b]
\centering
\includegraphics[width=\columnwidth]{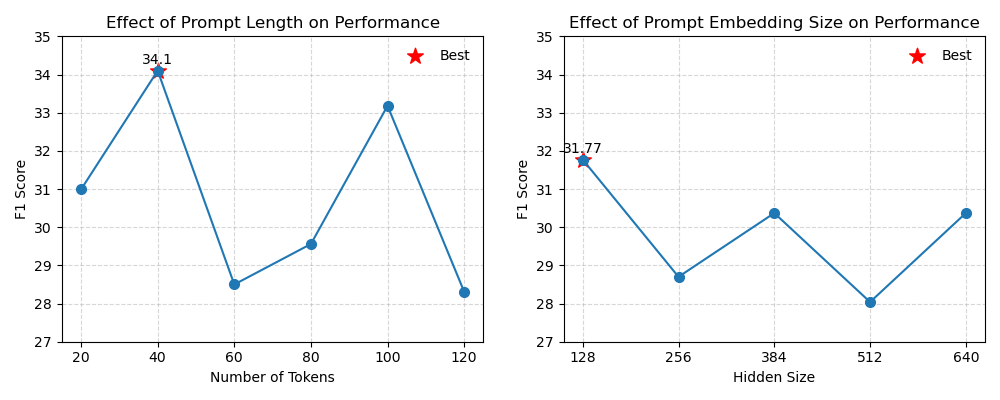}
\caption{Sensitivity of RCSP to soft prompt capacity using Gemma-3-12b-pt on TruthfulQA. Left: performance as a function of number of prompt tokens. Right: performance across different embedding dimensions.}
\label{fig:ablation-chart}
\end{figure}

Table~\ref{tab:ablation-truthfulqa-full} shows that disabling individual components of the objective leads to measurable degradation. Removing the curriculum schedule (i.e., fixed $\mu$) results in the largest increase in hallucination, while eliminating the contrastive loss noticeably reduces factual performance. These results confirm that the improvements observed  arise from the coordinated interaction of contrastive anti-supervision, utility preservation, curriculum scheduling and KL regularization rather than from instruction prompting alone.


We further analyze the sensitivity of RCSP to the number of soft prompt tokens using the Gemma-3 (12B) backbone on TruthfulQA by varying the prompt length while keeping all other training settings fixed. The results are shown in Figure \ref{fig:ablation-chart}, where we observe that peak performance is achieved at 40 soft prompt tokens. This suggests that while a minimum prompt capacity is necessary to encode the desired behavioral adjustments, excessively large prompts may not always be beneficial. We also analyze sensitivity to the soft prompt embedding dimension by varying the prompt hidden size while keeping all other training settings fixed. As shown in Figure \ref{fig:ablation-chart}, the best performance is obtained at hidden size 128, and increasing the embedding dimension did not lead to further gains in this case. In several cases, larger hidden sizes reduced both factual performance and overall F-score. Overall, these results indicate that RCSP can achieve strong performance with relatively small prompt sizes and embedding capacities, further supporting its efficiency as a lightweight adaptation method.


\section{Limitations and Future Work}
While RCSP demonstrates strong performance within a lightweight training framework, a few limitations and opportunities for further exploration remain. Although we rely on an LLM-as-a-Judge protocol for scalable evaluation, such approaches may introduce bias depending on the judge model and prompt formulation. To mitigate this, we adopt a strict grading rubric based on the SimpleQA framework and enforce deterministic decoding during evaluation to ensure consistent relative comparisons.

As described in Section 5.4, we report the F-score as the harmonic mean of Overall Accuracy (A) and Conditional Accuracy (C), following SimpleQA \cite{wei2024measuring}. The authors note that when model accuracy is below 50\%, it can still be rational for the model to guess when it is approximately 50\% confident in its response. While this metric provides a convenient performance summary, it aggregates factual responses, hallucinations and abstentions into a single value. Consequently, identical F-scores can arise from different behavioral profiles.  For example, a model that answers all questions with moderate accuracy and one that answers fewer questions with higher conditional accuracy may achieve the same F-score despite exhibiting different hallucination rates and abstention behavior. Future work could explore evaluation metrics that allow for deployment-specific cost weighting, which would enable more flexible alignment in different settings. 

Additionally, there is potential for further improvement by integrating prompt routing into this approach. Prompt routing has already shown promise in improving LLM performance in other applications \cite{choi2023smop}. Since soft prompts can function as a modular plug-and-play mechanism, different prompts can be dynamically selected based on deployment requirements to enhance reliability in specialized environments. For example, in security-critical applications, dynamic routing could apply a highly conservative soft prompt that strictly prioritizes abstention in more sensitive settings. We also note that methods that do not operate in the same efficiency paradigm or stage of the generation process might achieve stronger absolute performance, but do so under different computational and architectural assumptions and often do not take abstention awareness into account. RCSP could be complementary to these approaches and could possibly be combined with them, which we leave as future exploration. Evaluations could also be further extended by incorporating benchmarks beyond generative QA.

\section{Conclusion}
LLMs are increasingly used in security-sensitive settings, where hallucinated responses can introduce significant reliability risks. In this paper, we introduced Responsible Contrastive Soft Prompting (RCSP), a lightweight approach designed to mitigate hallucinations while encouraging responsible abstention in generative question-answering tasks. Rather than modifying backbone model parameters, RCSP trains learnable soft prompts to balance hallucination suppression with the preservation of factual utility.

This work investigated the three central research questions of our problem statement. Across multiple generative QA benchmarks, RCSP attains higher F-scores on the majority of datasets relative to the base LLM and instruction-based prompting baselines. These results demonstrate that RCSP can mitigate hallucinations without substantially degrading factual performance. Importantly, these improvements are obtained by training only a small number of parameters, resulting in a modular and computationally efficient design. Our findings demonstrate that RCSP provides affirmative answers to RQ1--RQ3 and shows potential as a practical and lightweight pathway for improving LLM reliability.

\begin{credits}
\subsubsection{Disclaimer} This paper identifies certain equipment, instruments, software, or materials to adequately describe the experimental procedure. Such identification is not intended to imply recommendation or endorsement of any product or service by NIST, nor is it intended to imply that the materials or equipment identified are necessarily the best available for the purpose.

\subsubsection{\ackname} The research reported herein was supported in part by NIST grant number 60NANB24D143. Any opinions, findings, conclusions, and recommendations expressed in this material are those of the author(s) and do not necessarily reflect the views of NIST. This work used Delta and DeltaAI at the National Center for Supercomputing Applications [award OAC 2005572] through allocation CIS251331 from the Advanced Cyberinfrastructure Coordination Ecosystem: Services and Support (ACCESS) program \cite{boerner2023access}, which is supported by U.S. National Science Foundation grants \#2138259, \#2138286, \#2138307, \#2137603, and \#2138296. The authors also acknowledge High Performance Computing at The University of Texas at Dallas (HPC@UTD) for providing computing resources.

\subsubsection{Disclosure of Interests.} The authors have no competing interests to declare that are relevant to the content of this article.

\end{credits}

%
%
\appendix
\section{Appendix: Further Experimental Settings}
\label{app:hyperparameters}

\noindent Due to space constraints, we only present below the addition we made to the grader prompt, adapted from the template provided in the SimpleQA benchmark \cite{wei2024measuring}, for our LLM-as-a-judge evaluation.

\lstset{
  basicstyle=\ttfamily\footnotesize,
  columns=fullflexible,
  breaklines=true,
  breakatwhitespace=true,
  keepspaces=true,
  showstringspaces=false
}
\begin{lstlisting} 
...consider the following predicted answers as correct: "Hyoong Won Choong", "Hyungwon Chung", or "Hyun Won Chung".

IMPORTANT:
- Use NOT_ATTEMPTED **only** when the answer explicitly expresses uncertainty (e.g., "I don't know", "not enough information", "cannot answer").
- Do **not** use NOT_ATTEMPTED for nonsensical, off-topic, or question-echoing answers (answers that merely repeat or rephrase the question without providing an answer); those should be graded INCORRECT.

Here is a new example....
\end{lstlisting}




In Table \ref{tab:hyperparams}, we also list the hyperparameter values used to train the soft prompts that we reported in our primary results table.

\begin{table}[!h]
\centering
\caption{Hyperparameter configuration for the soft prompts reported for Gemma-3-12b-pt and Llama-3.1-8B. Notation follows Equations (2) and (4) in the methodology section.}
\label{tab:hyperparams}
\begin{tabular}{lc@{\hspace{1.2em}}c}
\hline
\textbf{Hyperparameter} & \shortstack{\textbf{Gemma-3-}\\\textbf{12b-pt}} & \shortstack{\textbf{Llama-3.1-}\\\textbf{8B}} \\ \hline
Learning Rate & $3 \times 10^{-5}$ & $3 \times 10^{-5}$ \\
Training Batch Size & 32 & 32 \\
Number of Epochs & 3 & 4 \\
Weight Decay & 0.01 & 0.01 \\
Hallucination Penalty ($\lambda$) & 1 & 0.7 \\
Retain KL Scale ($\beta_{utility}$) & 3.8 & 7 \\
Suppress KL Scale ($\beta_{contrast}$) & 1 & 1.2 \\
Curriculum Start ($\mu_{start}$) & 0.4 & 0.4 \\
Curriculum End ($\mu_{end}$) & 0.9 & 1 \\
Warmup Start ($T_{start}$) & 0 & 0 \\
Warmup End ($T_{end}$) & 0.6 & 0.7 \\
Prompt Tokens & 40 & 128 \\
Prompt Hidden Size & 128 & 256 \\ \hline
\end{tabular}
\end{table}

\bibliographystyle{splncs04}
\bibliography{references}
%


\end{document}